\documentclass[]{article}
\usepackage[letterpaper]{geometry}
\usepackage{amta2016}
\usepackage{times}
\usepackage{url}
\usepackage{latexsym}
\usepackage{natbib}
\usepackage{layout}
\usepackage{algorithmic}
\usepackage{graphicx}

\begin{document}

\title{\bf Fast, Scalable Phrase-Based SMT Decoding}  

\author{\name{\bf Hieu Hoang} \hfill  \addr{hieu@moses-mt.org}\\ 
        \addr{Moses Machine Translation CIC, UK}
\AND
       \name{\bf Nikolay Bogoychev} \hfill \addr{s1031254@sms.ed.ac.uk}\\
        \addr{University of Edinburgh, Scotland}
\AND
       \name{\bf Lane Schwartz} \hfill \addr{lanes@illinois.edu}\\
        \addr{University of Illinois, USA}
\AND
       \name{\bf Marcin Junczys-Dowmunt} \hfill \addr{junczys@amu.edu.pl}\\
        \addr{Adam Mickiewicz University}
}

\maketitle
\pagestyle{empty}

\begin{abstract}
The utilization of statistical machine translation (SMT) has grown enormously over the last decade, many using open-source software developed by the NLP community. As commercial use has increased, there is need for  software that is optimized for commercial requirements, in particular, fast phrase-based decoding and more efficient utilization of modern multicore servers.

In this paper we re-examine the major components of phrase-based decoding and decoder implementation with particular emphasis on speed and scalability on multicore machines. The result is a drop-in replacement for the Moses decoder which is up to fifteen times faster and scales monotonically with the number of cores. 
\end{abstract}

\section{Introduction}

SMT has steadily progressed from a research discipline to commercial viability during the past decade as can be seen from services such as the Google and Microsoft Translation services. As well as general purpose services such as these, there is a large number of companies that offer customized translation systems, as well as  companies and organization that implement in-house solutions. Many of these customized solutions use Moses as their SMT engine.

For many users, decoding is the most time-critical part of the translation process. Making use of the multiple cores that are now ubiquitous in today's servers is a common strategy to ameliorate this issue. However, it has been noticed that the Moses decoder, amongst others, is unable to efficiently use multiple cores~\citep{mfernandez2016boosting}. That is, decoding speed does not substantially increase when more cores are used, in fact, it may actually \emph{decrease} when using more cores. There has been speculation on the causes of the inefficiency as well as potential remedies. 

This paper is the first we know of that focuses on improving decoding speed on multicore servers. We take a holistic approach to solving this issue, creating a decoder that is optimized for multi-core processing speed by concentrating on four main areas:
\begin{enumerate}
  \item \vspace{-2 mm} Faster memory management of data-structures through the use of customized memory pools
  \item \vspace{-2 mm} Exploring alternatives to cardinality-based hypothesis stack configuration
  \item \vspace{-2 mm} Re-examining the efficiency of phrase-table lookup using translation rule caching and data compression
  \item \vspace{-2 mm} Integrating the lexicalized re-ordering model into the phrase-table, thus eliminating the need for independent random lookup this model 
\end{enumerate}

The result is a decoder that is significantly faster than the Moses baseline for single-threaded operation, and scales with the number of cores.

We will maintain the Moses decoder's embarrassingly parallel, one sentence-per-thread decoding framework. As far as possible, model scores and functionality are compatible with Moses to aid comparison and ease transition for existing users. The source code is available in the existing Moses repository\footnote{\scriptsize https://github.com/moses-smt/mosesdecoder/tree/master/contrib/moses2}.

The rest of the paper will be broken up into the following sections. The rest of this section will discuss prior work and an outline of the phrase-based model. Section~\ref{sec:Proposed Improvements} will then describe the modifications to improve decoding speed. We describe the experiment setup in Section~\ref{sec:Experimental Setup} and present results Section~\ref{sec:Results}. We conclude in the last section and discuss possible future work.

\subsection{Prior Work}

Most prior work on increasing decoding speed look to optimizing specific components of the decoder or the decoding algorithm. 

~\cite{Heafield-kenlm} and ~\cite{yasuhara-EtAl:2013:EMNLP} describe fast, efficient data structures for language models. ~\cite{zens2007efficient} describes an implementation of a phrase-table for an SMT decoder that is loaded on demand, reducing the initial loading time and memory requirements. ~\cite{junczys_tsd_2012b} extends this by compressing the on-disk phrase table and lexicalized re-ordering model. 

~\cite{Chiang:2007:cl} describes the cube-pruning and cube-growing algorithm which allows the tradeoff between speed and translation quality to the adjusted with a single parameter. ~\cite{wuebker2012fast} note that language model querying is amongst the most expensive operation in decoding. They sought to improved decoding speed by caching score computations early pruning of translation options. This work is similar to ~\cite{Heafield-mtplz} which group hypotheses with identical language model context and incrementally expand them, reducing LM querying.

~\cite{mfernandez2016boosting} was concerned with multi-core speed but treated decoding as a black box within a parallelization framework.

There are a number of phrase-based decoding implementations, many of which implements the extensions to the phrase-based model described above. The most well known is Moses~\citep{koehn-EtAl:2007:PosterDemo} which implements a number of speed optimizations, including cube-pruning. It is widely used for MT research and commercial use. 

Joshua~\citep{Joshua-Decoder} also supports cube-pruning for phrase-based models. Phrasal~\citep{spence2014phrasal} supports a number of variants of the phrase-based model. Jane~\citep{peitz2012jane} supports the language model look-ahead described in ~\cite{wuebker2012fast} in addition to other tools to speed up decoding such as having a separate fast, lightweight decoder. mtplz is a specialized decoder developed to implement the incremental decoding described in ~\cite{Heafield-mtplz}.

The Moses, Joshua and Phrasal decoders implement multithreading, however, they all report scalability problems, either in the paper (Phrasal) or via social media (Moses\footnote{\scriptsize https://github.com/moses-smt/mosesdecoder/issues/39} and Joshua\footnote{\scriptsize https://twitter.com/ApacheJoshua/status/342022794097340416}).

Jane and mtplz are single-threaded decoders, relying on external applications to parallelize operations. 

This paper not only focuses on faster single-threaded decoding but also on overcoming the shortcomings of existing decoding implementations on multicore servers. Unlike ~\cite{mfernandez2016boosting}, we will optimize decoding speed by looking inside the black box. We will compare multicore performance the best-of-breed phrase-table described in ~\cite{junczys_tsd_2012b} with our own implementation. We will use the cube-pruning algorithm, however, the standard phrase-based decoding algorithm is also available and a framework exists to accommodate other decoding algorithms in future. We use KenLM~\citep{Heafield-kenlm} due to its popularity and consistent performance, but as with Moses, other language model implementations can be added later.

\subsection{Phrase-Based Model}

The objective of decoding is to find the target translation with the maximum probability, given a source sentence. That is, for a source sentence $s$, the objective is to find a target translation $\hat{t}$ which has the highest conditional probability $p(t | s)$. Formally, this is written as:
\begin{equation}
\hat{t} = \arg \max_t p( t | s )
\label{eq:argmax-factored-trans}
\end{equation}
where the \emph{arg max} function is the search. The log-linear model generalizes Equation~\ref{eq:argmax-factored-trans} to include more component models and weighting each model according to the contribution of each model to the total probability. 
\begin{equation}
\label{eq:Log-Linear}
p(t | s) 	=  \frac{1}{Z} \exp ( \sum_m \lambda_m h_m ( t, s)^{} )
\end{equation}
where $\lambda_m$ is the weight, and $h_m$ is the \emph{feature function}, or `score', for model $m$. $Z$ is the partition function which can be ignored for optimization. 

The standard feature functions in the phrase-based model include:
\begin{enumerate}
  \item \vspace{-2 mm} a distortion penalty
  \item \vspace{-2 mm} a phrase-penalty,
  \item \vspace{-2 mm} a word penalty,
  \item \vspace{-2 mm} an unknown word penalty.
  \item \vspace{-2 mm} log transforms of the target language model probability $p(t)$, 
  \item \vspace{-2 mm} log transforms translation model probabilities, $p_{TM}(t|s) $ and $p_{TM}(s|t)$, and word-based translation probabilities $p_w(t|s)$ and $p_w(s|t)$,
  \item \vspace{-2 mm} log transforms of the lexicalized re-ordering probabilities,
\end{enumerate}

Of these feature functions, we will focus on optimizing the speed of the phrase-table and lexicalized re-ordering models.


\subsection{Beam Search}

A translation of a source sentence is created by applying a series of translation rules which together translate each source word once, and only once. Each partial translation is known as a \emph{hypothesis}, which is created by applying a rule to an existing hypothesis. This \emph{hypothesis expansion} process starts with a hypothesis that has translated no source word and ends with completed hypotheses that has translated all source words. The highest-scoring completed hypothesis, according to the model score, is considered the best translation, $\hat{t} $.

In the phrase-based model, each rule translates a contiguous sequence of source words. Successive applications of translation rules do not have to be adjacent on the source side, depending on the distortion limit. The target output is constructed strictly left-to-right from the target side using this series of translation rules. 

A beam search algorithm is used to create the completed hypothesis set efficiently. Hypotheses are grouped into stacks where each stack holds a number of comparable hypotheses. Most phrase-based implementations group hypotheses according to coverage cardinality. 

\section{Proposed Improvements}
\label{sec:Proposed Improvements}

We will also concentrate on four main areas for optimization.

\subsection{Efficient Memory Allocation}

The search algorithm creates and destroy a large number of intermediate objects such as hypotheses and feature function states. This puts a burden on the operating system due to the need to synchronize memory access, especially when using a large number of threads. Libraries such as tcmalloc~\citep{ghemawat2009tcmalloc} are designed to reduce locking contention for multi-threaded application but in our case, this is still not enough. 

We shall seek to improve decoding speed by replacing the operating system's general purpose memory management with our own custom memory management scheme. Memory will be allocated from a memory pool rather than use the operating system's general purpose allocation functions.

A memory pool is a large block of memory that has been given to the application by the operating system. The application is then responsible for allocating portions of this memory to its components when requested. We will use thread-specific memory pools to increase speed by avoiding locking contention during memory access. Our memory pools will be dynamic. That is, the memory requirement does not have to be known or specified before running the application, the pool can grow when required but they will never reduce in size. The pools are deleted only when the application ends. 

To further increase memory management speed, objects in the memory pool are not deleted. Unused data structures accumulates in the pool until a reset event. The pool is assumed to be empty and simply reused after the event. We instantiate two memory pools per decoding thread, one which is never reset and another which is reset after the decoding of each sentence. Data structures are created in either pool according to their life cycle.

Accumulating unused objects in the memory pools can result in unacceptably high memory usage so object queues are available for high-churn objects which allows the decoder to re-cycle unused objects before the reset event. This also increases speed as LIFO queues are used so that the most recently accessed memory are used, increasing CPU cache hits.



\subsection{Stack Configurations}

The most popular stack configuration for phrase-based models is coverage cardinality, that is, hypotheses that have translated the same number of source words are stored in the same stack. This is implemented in Pharaoh, Moses and Joshua.


However, there are alternatives to this configuration. ~\cite{Och:2001a} uses a single stack for all hypotheses, ~\cite{Brown:1993} uses coverage stacks (ie. one stack per unique coverage vector) while ~\cite{peitz2012jane} and ~\cite{Zens+Ney:2008:iwslt} apply both coverage and cardinality pruning. While useful, these prior works present only one particular stack configuration each. ~\cite{ortizmartinez-garciavarea-casacuberta:2006:WMT} explore a range of stack configurations by defining a granularity parameter which controls the maximum number of stacks required to decode a sentence. 

We shall re-visit the question of stack configuration with a particular emphasis on how they can help improve the tradeoff between speed and translation quality. We will do so in the context of the cube-pruning algorithm, the algorithm that we will be using and which was not available to many of the earlier work.

\subsection{Phrase-Table Optimizations}

\begin{figure}
\centering
\begin{tabular}{cc}
{\includegraphics[scale=0.4]{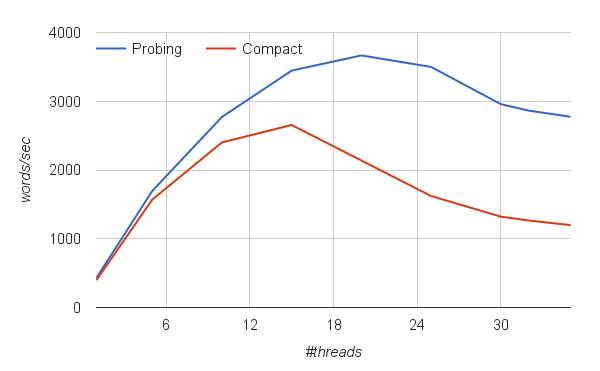}} 
\end{tabular}
\caption{Moses decoding speed with two different phrase-table implementations}
\label{fig:moses-phrase-tables-time}
\end{figure}

For any phrase-table table of a realistic size, memory and loading time constraints requires us to use a load-on-demand implementation. Moses has several which we can make use of, each with differing performance characteristics. Figure~\ref{fig:moses-phrase-tables-time} shows the decoding speed for the fastest two implementations.
From this, it appears that the Probing phrase-table~\citep{Bogoychev.probing.2016} has the fastest translation rule lookup, especially with large number of cores, therefore, we will concentrate exclusively on this implementation from hereon.

We propose two optimizations. Firstly, the translation rule caching mechanism in Moses saves the most recently used rules. However, this require locking and active management in clearing of old rules. The result is \emph{slower} decoding, Table~\ref{tab:phrase-table-cache}. 
\begin{table}
\small
\begin{center}
\begin{tabular}{|l|c|c|} \hline
		& No cache	& Caching \\ \hline
Decoding time  	& 2877	& 2540 (-12\%) \\ \hline
\end{tabular}
\end{center}
\caption{Decoding speed (in words / sec with 32 threads) when using phrase-table cache}
\label{tab:phrase-table-cache}
\end{table}

We shall explore a simpler caching mechanism by creating a static cache of the most likely translation rules to be used at the start of decoding.

Secondly, the Probing phrase-table use a simple compression algorithm to compress the target side of the translation rule. Compression was championed by ~\cite{junczys_tsd_2012b} as the main reason behind the speed of their phrase-table but as we saw in Figure~\ref{fig:moses-phrase-tables-time}, this comes at the cost of scalability to large number of threads. We shall therefore take the opposite approach to and improve decoding speed by disabling compression.

\subsection{Lexicalized Re-ordering Model Optimizations}

Similar to the phrase-table, the lexicalized re-ordering model is trained on parallel data. A resultant model file is then queried during decoding. The need for random lookup during querying inevitably results in slower decoding speed. Previous work such as~\cite{junczys_tsd_2012b} improve querying speed with more compact data structures.

However, the model's query keys are the source and target phrase of each translation rule. Rather than storing the lexicalized re-ordering model separately, we shall integrating it into the translation model, eliminating the need to query a separate file. However, the model remains the same under the log-linear framework, including having its own weights. 

This optimization has precedent in ~\cite{peitz2012jane} but the effect on decoding speed were not published. We will compare results with using a separate model in this paper.

\section{Experimental Setup}
\label{sec:Experimental Setup}

We trained a phrase-based system using the Moses toolkit with standard settings. The training data consisted of most of the publicly available Arabic-English data from Opus~\citep{tiedemann2012parallel} containing over 69 million parallel sentences, and tuned on a held out set. The phrase-table was then pruned, keeping only the top 100 entries per source phrase, according to $p(t|s)$. All model files were then binarized; the language models were binarized using KenLM~\citep{Heafield-kenlm}, the phrase table using the Probing phrase-table, lexicalized re-ordering model using the compact data structure~\citep{junczys_tsd_2012b}. These binary formats were chosen for their best-in-class multithreaded performance. Table~\ref{tab:model-files} gives details of the resultant sizes of the model files. For testing decoding speed, we used a subset of the training data, Table~\ref{tab:test-sets}. 
\begin{table}
\begin{center}
\begin{tabular}{|l|c|c|} \hline
		& ar-en	& fr-en \\ \hline
Phrase table  	& 17 	& 5.8 \\
Language model (5-gram) 	& 3.1  	& 1.8 \\ 
Lex re. model	& 2.3	& 0.6 \\ \hline
\end{tabular}
\end{center}
\caption{Model sizes in GB}
\label{tab:model-files}
\end{table}

For verification with a different dataset, we also used a second system trained on the French-English Europarl corpus (2m parallel sentences). The two different systems have characterics that we are interested in analyzing; ar-en have short sentences with large models while fr-en have overly long sentences with smaller models. Where we need to compare model scores, we used held out test sets.

\begin{table}
\begin{center}
\small
\begin{tabular}{|l|r|r|} \hline
		& ar-en		& fr-en \\ \hline
\multicolumn{3}{|l|}{For speed testing} \\ \hline
Set name	& \multicolumn{2}{|c|}{Subset of training data} \\
\# sentences  	& 800k 		& 200k \\
\# words 	& 5.8m 		& 5.9m \\ 
Avg words/sent	& 7.3		& 29.7 \\ \hline
\multicolumn{3}{|l|}{For model score testing} \\ \hline
Set name	& OpenSubtitles	& newstest2011 \\
\# sentences  	& 2000 		& 3003 \\
\# words 	& 14,620 	& 86,162 \\ 
Avg words/sent	& 7.3		& 28.7 \\ \hline
\end{tabular}
\end{center}
\caption{Test sets}
\label{tab:test-sets}
\end{table}

Standard Moses phrase-based configurations are used, except that we use the cube-pruning algorithm~\citep{Chiang:2007:cl} with a pop-limit of 400\footnote{the pop-limit was chosen from public discussion on the Moses mailing list on an acceptable balance between decoding speed and translation quality with Moses for commercial use 
}, rather than the basic phrase-based algorithm. The cube-pruning algorithm is often employed by users who require fast decoding as it gives them the ability to trade speed with translation quality via a simple pop-limit parameter.

As a baseline, we use a recent\footnote{The experiments were performed between January and May 2016 with the latest github code to hand. The main ar-en experiments were rerun with the source code as of 8th June, 2016 to ensure there were no material difference. The commit hash was \scriptsize{bc5f8d15c6ce4bc678ba992860bfd4be6719cee8} } version of the Moses decoder taken from the github repository.

For all experiments, we used a Dell PowerEdge R620 server with 16 cores, 32 hyper-threads, split over 2 physical processors (Intel Xeon E5-2650 @ 2.00GHz). The server has 380GB RAM. The operating system was Ubuntu 14.04, the code was compiled with gcc 4.8.4 and Boost 1.59\footnote{http://boost.org/} and the tcmalloc library.

\section{Results}
\label{sec:Results}

\subsection{Optimizing Memory}


Over 24\% of the Moses decoder running time is spent on memory management (Table~\ref{tab:optimizing-memory}). This increases to 39\% when 32 threads are used, dampening the scalability of the decoder. By contrast, our decoder spends 11\% on memory management and does not significantly increase with more threads.
\begin{table}
\begin{center}
\small
\begin{tabular}{|l|c|c|c|c|} \hline
		& \multicolumn{2}{c|} {Moses}	& \multicolumn{2}{c|} {Our Work} \\ \hline
\# threads	& 1 		& 32	& 1 		& 32  \\ \hline
Memory  	& 24\%		& 39\% 	& 11\%		& 13\% \\
LM 		& 12\%	 	& 2\% 	& 47\%		& 38\% \\ 
Phrase-table	& 9\%	 	& 5\% 	& 2\%		& 4\% \\ 
Lex RO 		& 8\%	 	& 2\% 	& 2\%		& 2\% \\ 
Search 		& 2\%	 	& 0\% 	& 14\%		& 19\% \\ 
Misc/Unknown	& 45\%	 	& 39\% 	& 24\%		& 29\% \\ \hline
\end{tabular}
\end{center}
\caption{Profile of \%age decoding time}
\label{tab:optimizing-memory}
\end{table}

\begin{figure}
\centering
\begin{tabular}{cc}
{\includegraphics[scale=0.4]{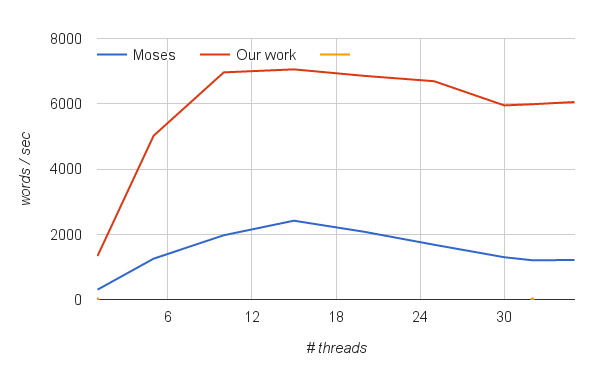}} 
\end{tabular}
\caption{Decoding speed of Moses and our decoder, using the same models}
\label{fig:mempool}
\end{figure}

Figure~\ref{fig:mempool} compares the decoding speed for Moses and our decoder, using the same models, parameters and test set. Our decoder is 4.4 times faster with one thread, and 5.0 times faster using all cores. Like Moses, however, performance actually worsens after approximately 15 threads. 


\subsection{Stack Configuration}

We investigated the effects of the following three stack configurations on model score and decoding speed:
\begin{enumerate}
  \item \vspace{-2 mm} coverage cardinality,
  \item \vspace{-2 mm} coverage,
  \item \vspace{-2 mm} coverage and end position of most recently translated source word.
\end{enumerate}
Coverage cardinality is the same as that in Moses and Joshua. Coverage configuration uses one stack per unique coverage vector. \emph{Coverage and end position} of most recently translated source word extends the coverage configuration by separating hypotheses where the position of the last translate word are different, even if the coverages are identical. 

This is an optimization to reduce the number of checks on the distortion limit, which is dependent on the last word position. The check is a binary function $d(C_{h}, e_{hypo}, range_r)$, where $C_{h}$ is the coverage vector of hypothesis $h$, $e_{h}$ is the end position of most recent source word that has been translated, and $range_r$ is the coverage of the rule to be applied. 

By grouping hypotheses according to coverage \emph{and} end position, the distortion limit only needs to be checked for each group. 
However, stack pruning occurs on each hypothesis group independently, therefore, potentially affecting search errors and model scores. 

\begin{figure}
\centering
\begin{tabular}{cc}
{\includegraphics[scale=0.4]{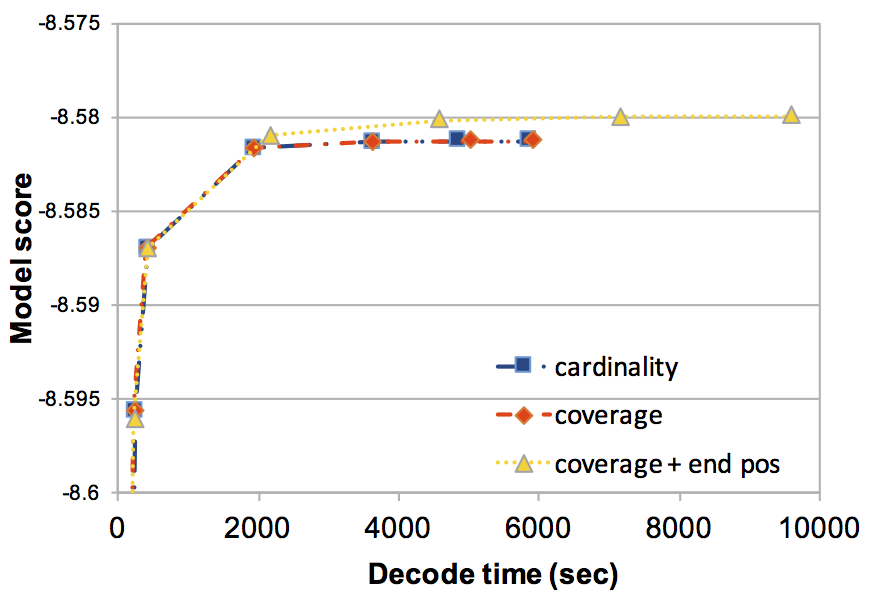}} 
\end{tabular}
\caption{Trade-off between decoding time average model scores for different stack configurations}
\label{fig:stack-configuration}
\end{figure}

Figure~\ref{fig:stack-configuration} presents the tradeoff between decoding time and average model scores, created by varying the cube-pruning pop-limit. None of the different stack configurations significantly outperform the others in either quality or decoding speed. However, the coverage \& end position produces slightly higher model scores at higher pop-limits, therefore, we continue to use this configuration throughout the rest of this paper.



\begin{figure}
\centering
\begin{tabular}{cc}
{\includegraphics[scale=0.5]{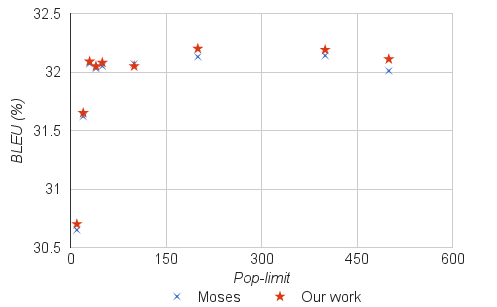}} 
\end{tabular}
\caption{Translation quality for different pop-limits}
\label{fig:bleu}
\end{figure}

We verified that the translation quality of our decoder is comparable to that of Moses in Figure~\ref{fig:bleu}, given the same parameters and models. This fits in with our intention of creating a drop-in replacement for the Moses decoder. 

\subsection{Translation Model}

In the first optimization, we create a static translation model cache containing translation rules that translates the most common source phrases. This is constructed during phrase-table training based on the source counts. The cache is then loaded when the decoder is started. It does not require the overhead of managing an active cache but there is still some overhead in using a cache. 
\begin{table}
\small
\begin{center}
\begin{tabular}{|r|r|r|} \hline
Cache size	& Decoding Time & Cache Hit \%age	\\ \hline
Before caching	& 229	& N/A	 \\
0		& 239 (+4.4\%)	& 0\% \\
1,000		& 213 (-7.0\%)	& 11\% \\
2,000		& 204 (-10.9\%)	& 13\% \\
4,000		& 205 (-10.5\%)	& 14\% \\
10,000		& 207 (-9.7\%)	& 17\% \\ \hline
\end{tabular}
\end{center}
\caption{Decoding time (in secs with 32 threads) for varying cache sizes}
\label{tab:cache}
\end{table}
Overall however, using a static cache result in a 10\% decrease in decoding time if the optimum cache size is used, Table~\ref{tab:cache}.

For the second optimization, we disable the compression of the target side of the translation rules. This increase the size of the binary files from 17GB to 23GB but the time saved not needing to decompress the data resulted in a 1.5\% decrease in decoding time with 1 thread and nearly 7\% when the CPUs are saturated, Table~\ref{tab:compression}.
\begin{table}
\small
\begin{center}
\begin{tabular}{|r|r|r|} \hline
\# threads	& Compressed pt & Non-compressed pt \\ \hline
1		& 3052	& 3006 (-1.5\%) \\
5		& 756	& 644 (-14.8\%) \\
10		& 372	& 362 (-2.7\%) \\
15		& 284	& 250 (-12.0\%) \\
20		& 244	& 227 (-7.0\%) \\
25		& 218	& 209 (-4.1\%) \\
30		& 206	& 192 (-6.8\%) \\
35		& 203	& 189 (-6.9\%) \\ \hline
\end{tabular}
\end{center}
\caption{Decoding time (in secs with 32 threads) for compressed and non-compressed phrase-tables}
\label{tab:compression}
\end{table}

\subsection{Lexicalized Re-ordering Model}

The lexicalized re-ordering model requires a probability distribution of the re-ordering behaviour of each translation rule learnt from the training data. This is represented in the model file as a fixed number of probabilities for each rule, exactly how many probabilities is dependant on the model's parameterization during training. During decoding, a probability from this distribution is assigned to each hypothesis according to the re-ordering of the translation rule.

Rather than holding the model probability distributions in the separate file, we pre-process the translation model file to include the lexicalized re-ordering model distributions for each rule. During decoding, the probability distribution is then taken from the translation model instead of querying a separate file. 

\begin{figure}
\centering
\begin{tabular}{cc}
{\includegraphics[scale=0.4]{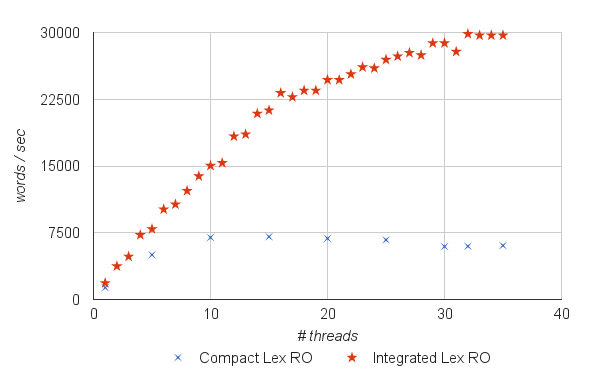}} 
\end{tabular}
\caption{Decoding speed with Compact Lexicalized Re-ordering, and integrated into a model the phrase-table}
\label{fig:lex-ro}
\end{figure}

This resulted in a significant decrease in decoding time, especially with high number of cores, Figure~\ref{fig:lex-ro}. 
Decoding speed increased by 40\% when using one thread but is 5 times faster when using 32 threads. 

\subsection{Scalability}

\begin{figure}
\centering
\begin{tabular}{cc}
{\includegraphics[scale=0.4]{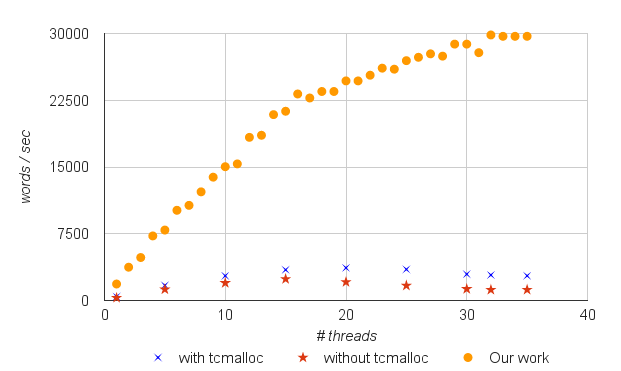}} 
\end{tabular}
\caption{Comparison of decoding speed of our work and Moses (with \& without the tcmalloc library}
\label{fig:speed}
\end{figure} 

Figure~\ref{fig:speed} shows decoding speed against the number of threads used. In our work, there is a constant increase in decoding speed when more threads are used, decreasing slightly after 16 threads when virtual cores are employed by the CPU. Overall, decoding is 16 times faster than single-threaded decoding when all 16 cores (32 hyperthreads) are fully utilized.


This contrast with Moses where speed increases up to approximately 16 threads but then become slower thereafter. Using the tcmalloc library has a small positive effect on decoding speed but does little to improve scalability 


Our work is 4.3 times faster than Moses with a single-thread and 10.4 faster when all cores are used.

 
\subsection{Other Models and Even More Cores}

\begin{figure}
\centering
\begin{tabular}{cc}
{\includegraphics[scale=0.4]{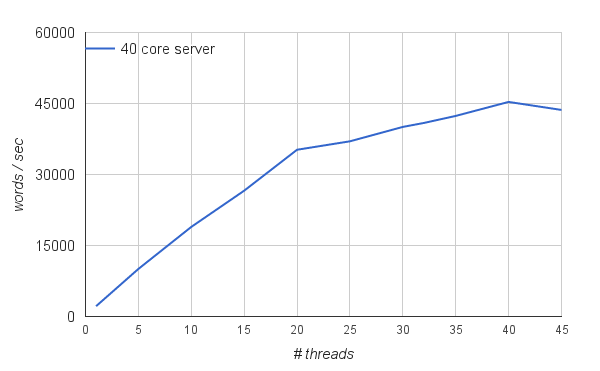}} 
\end{tabular}
\caption{Decoding speed of our decoder with more cores}
\label{fig:more-cores}
\end{figure}

Our decoder show no scalability issues when we tested with the same model and tested set on a larger server, Figure~\ref{fig:more-cores}.

\begin{figure}
\centering
\begin{tabular}{cc}
{\includegraphics[scale=0.4]{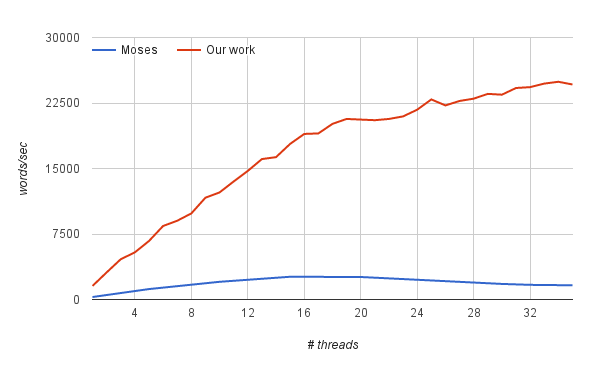}} 
\end{tabular}
\caption{Decoding speed for fr-en model}
\label{fig:fr-en-speed}
\end{figure}

We verify the results with the French-English phrase-based system and test set. The speed gains are even greater than the Arabic-English test scenario, Figure~\ref{fig:fr-en-speed}. Our decoder is 5.4 times faster than Moses with a single-thread and 14.5 faster when all cores are saturated.

\begin{figure}
\centering
\begin{tabular}{cc}
{\includegraphics[scale=0.4]{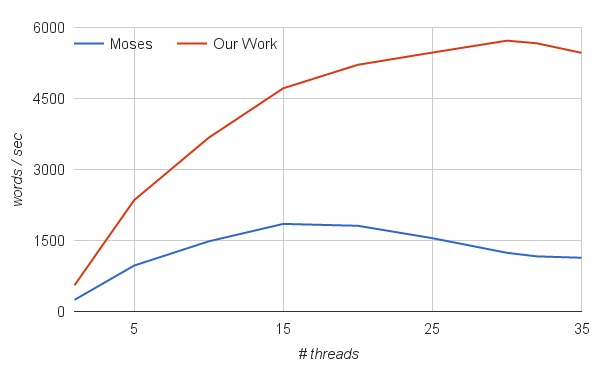}} 
\end{tabular}
\caption{Decoding speed with a large language model}
\label{fig:large-lm}
\end{figure}

It has been suggested that using a larger language model would overpower the improvements in decoding speed. We tested this conjecture by replacing the language model in the ar-en experiment with a 96GB language model. The time to load of language model is significant (394 sec) and was excluded from the translation speed. Results show that our decoder is 7 times faster than Moses and still scales monotonically until all CPUs are saturated, Figure~\ref{fig:large-lm}.

\section{Conclusion}

We have presented a new decoder that is compatible with Moses. By studying the shortcomings of the current implementation, we are able to optimize for speed, particularly for multicore operation. This resulted in double digit gains compared to Moses on the same hardware. Our implementation is also unaffected by scalability issues that have afflicted Moses. 

In future, we shall investigate other major components of the decoding algorithm, particularly the language model which has not been touched in this paper. We are also keen to explore the underlying reasons for the scalability issues in Moses to get a better understanding where potential performance issues can arise.

 \section*{Acknowledgments}
This work is sponsored by the Air Force Research Laboratory, prime contract FA8650-11-C-6160.  The views and conclusions contained in this document are those of the authors and should not be interpreted as representative of the official policies, either expressed or implied, of the Air Force Research Laboratory or the U.S. Government.

Thanks to Kenneth Heafield for advice and code.
\small

\bibliographystyle{apalike}
\bibliography{amta2016,mt,more}

\end{document}